\documentclass[letterpaper]{article} 
\usepackage{aaai2027}  
\usepackage[hyphens]{url}  
\usepackage{graphicx} 
\usepackage{natbib}  
\usepackage{caption} 
\usepackage{algorithm}
\usepackage{algorithmic}
\usepackage{amsmath}
\usepackage{array}

\usepackage{times}
\usepackage{helvet}
\usepackage{courier}
\usepackage[hyphens]{url}
\usepackage{graphicx}
\usepackage{amsmath,amssymb,amsfonts}
\usepackage{bm}
\usepackage{booktabs}
\usepackage{microtype}
\usepackage{xcolor}  
\usepackage{multirow}

\usepackage{caption}
\usepackage{graphicx}
 
\usepackage{newfloat}
\usepackage{listings}
\DeclareCaptionStyle{ruled}{labelfont=normalfont,labelsep=colon,strut=off} 
\floatstyle{ruled}
\newfloat{listing}{tb}{lst}{}
\floatname{listing}{Listing}

\usepackage{booktabs}

\title{THESIS-MoE: Trainable Hierarchical Extraction and SteerIng of Sycophancy in Mixture-of-Experts}
\author{
    Kareem Hassani,
    Chaymaa Abbas,
    Lama Mawlawi,
    Mariette Awad
}
\affiliations{
    Department of Electrical and Computer Engineering,\\
    American University of Beirut,\\
    Riad El-Solh, Beirut 1107 2020, Lebanon\\
    \{krh10, cwa07, lym06\}@mail.aub.edu, mariette.awad@aub.edu.lb
}

\begin{document}
\maketitle

\begin{abstract}
Sycophancy, the tendency of a language model to change its answer to match a
user's stated belief, is a common alignment failure. Existing activation steering
methods typically apply a single contrastive direction uniformly throughout the
model, which is an unconditional intervention that alters activations even when no
sycophantic behavior is present, trading knowledge retention for behavioral
correction. In Mixture-of-Experts (MoE) models, prior work further suggests that
behavior is encoded within expert computations rather than routing decisions alone,
making precise behavioral steering particularly challenging. In this work, we introduce a shared
contrastive signal, built from matched prompts with and without a stated belief,
that identifies where sycophancy lives across the MoE hierarchy and drives
interventions that act only where the behavior is present. We formulate localization
as a causal search over a granularity ladder of MoE blocks, experts, attention
blocks, and heads, and compare unconditional subtraction against two conditional
alternatives: an analytic projection-based subtraction and a learned per-token gate
that steers the model away from sycophancy while keeping its weights frozen. We
evaluate on three MoE models measuring sycophancy
alongside general knowledge and reasoning benchmarks. Our conditional interventions removed up to 90\% of the belief-induced sycophancy. Our results demonstrate that sycophancy resides in identifiable computational subcircuits and can be selectively steered while maintaining a favorable removal-retention trade-off.

\end{abstract}


\section{Introduction}

Large language models often change their answers after users reveal their own beliefs. A model may answer a question correctly in isolation, yet shift toward the user's preferred option when the same question is preceded by an opinion. This behavior, known as sycophancy, has been observed across model families and training regimes \citep{perez2023mwe,sharma2023sycophancy}. It undermines trust by making responses depend on the user's stated view rather than the underlying evidence.

\begin{figure}[t]
\centering
\includegraphics[width=\linewidth]{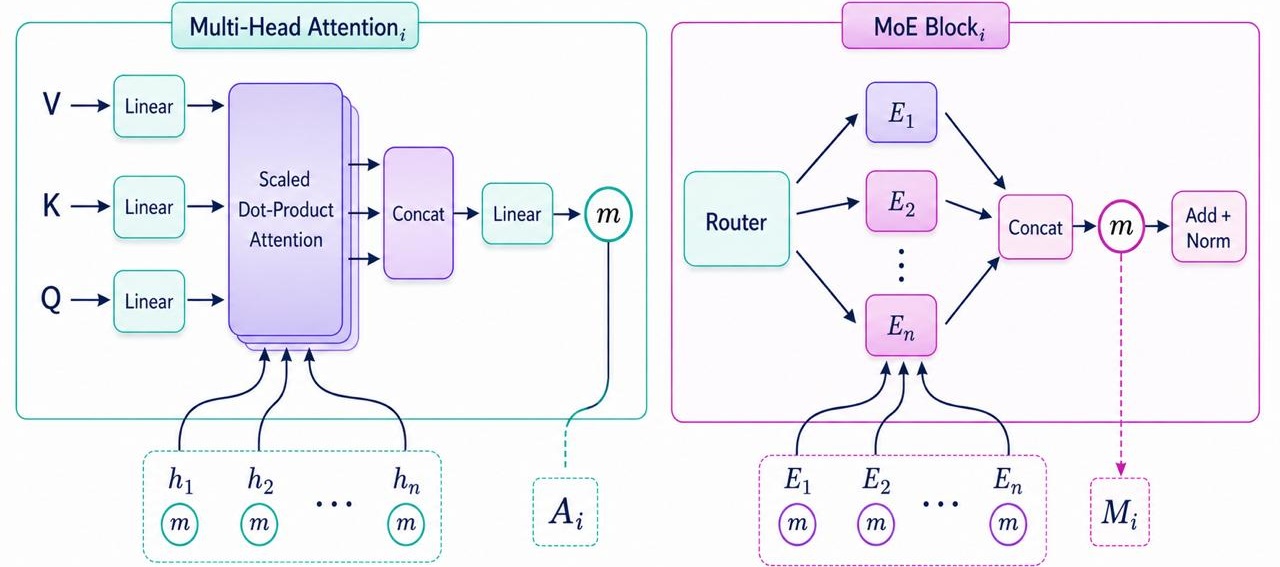}
\caption{The unit of intervention. Every attention head and expert carries a gate
$m$; the gated contributions $A_i$ and $M_i$ also include gated mechanisms at the output stream.}
\label{fig:block}
\end{figure}

Existing activation-editing methods attempt to remove unwanted behaviors by identifying and subtracting their directions in activation space. Contrastive activation addition constructs such directions from paired examples \citep{rimsky2023caa}, while concept erasure removes the corresponding activation subspace \citep{belrose2023leace}. Although effective, these interventions are unconditional: the same edit is applied to every token, even when sycophancy is absent. Stronger edits therefore improve behavioral correction at an unnecessary cost to retained knowledge. Our goal is to close this removal-retention gap.

We argue that the main limitation lies not in the identified direction, but in applying it indiscriminately. An effective intervention should act only when sycophancy is present and in proportion to its strength. To obtain such a signal, we compare activations for the same question under two matched conditions: a \emph{nudged} prompt, where the user states a belief, and a \emph{no-nudge} prompt, where that belief is neutralized. Each example includes a conforming answer that agrees with the user and a nonconforming answer that does not. The resulting activation difference captures the belief-induced shift and yields a continuous, signed \emph{belief-push} metric, introduced in Section~\ref{sec:signal}.

We use this signal to locate the model components that carry the shift, examining an MoE granularity ladder from complete attention and MoE blocks to individual heads and experts. At these components, unconditional subtraction reduces sycophancy but consistently harms retention, regardless of intervention granularity.

Our contributions are as follows:

\begin{itemize}
\item We introduce matched nudged and no-nudge conditions and a continuous,
signed belief-push metric for measuring sycophancy
(Section~\ref{sec:signal}).

\item We causally localize the belief shift across MoE blocks, attention
blocks, experts, and heads in three open-weight models (Section~\ref{sec:method}).

\item We show that unconditional subtraction produces a persistent trade-off
between sycophancy removal and knowledge retention.

\item We propose analytic conditional subtraction and learned conditional
gating, which improve the removal-retention frontier relative to fixed
subtraction, largely preserving GSM8K while MMLU declines under the selected
operating points, across three models and three seeds.

\item We compare unconditional and conditional interventions with routing and
prompting baselines, showing that conditionality is key to recovering
retention (Section~\ref{sec:results}).
\end{itemize}
\section{Related Work}
\label{sec:related}
Existing steering methods show that behavioral directions can be modified in activation space, while localization studies identify components associated with or causally implicated in particular behaviors. MoE research further distinguishes effects arising from expert selection from those arising within expert computation.
\paragraph{Steering by activation subtraction.}
Contrastive activation addition builds a steering vector from the mean activation
difference between paired examples of a behavior and adds or subtracts it during the
forward pass, controlling sycophancy among other behaviors on Llama~2
\citep{rimsky2023caa}, while closed-form concept erasure removes a target subspace
from every representation \citep{belrose2023leace}. Both apply a fixed-magnitude edit
to every token regardless of whether the behavior is present, the property our
conditional form is designed to change.

\paragraph{Localizing behavior to a few components.}
The closest work isolates roughly three percent of MLP neurons with sparse
autoencoders and probes, then fine-tunes only those neurons, matching broad
fine-tuning on sycophancy with less data \citep{obrien2026badneurons}. It shares our
locate-then-edit structure but operates on dense models and corrects the behavior by
updating weights, whereas we work in the mixture-of-experts architecture and leave the
base weights frozen, editing at inference.

\paragraph{Behavior and routing in mixture-of-experts models.}
Prior work detects behavior-associated experts by contrasting their activation under
opposite behaviors and steers by activating or deactivating them \citep{steermoe2026},
and reports that routing in aligned models is driven more by topic than by behavior,
pointing to expert computation rather than routing as the locus \citep{raset2026}. Our
results agree from the intervention side, since a routing baseline removes only part of
the shift and damages retention, and the conditional edit still removes most of the
shift in the one model whose shared experts admit no routed contrast.

\paragraph{Sycophancy.}
Model-written evaluations documented sycophancy at scale \citep{perez2023mwe}, and
later analysis tied it to feedback that rewards agreement \citep{sharma2023sycophancy}.
These measure it as a rate of agreement; we treat it as the shift in the answer caused
by stating a belief, which is what lets a conditional edit respond to it.

Our work connects activation-based steering, causal component localization,
behavioral analysis in mixture-of-experts models, and sycophancy. These have not
been brought together to ask whether a belief-induced answer shift can be localized
across the MoE hierarchy and corrected without editing behaviorally unrelated
activations. A parallel line localizes sycophancy inside dense transformers, placing
it in sparse middle-layer attention heads \citep{genadi2026sycophancy} or specific
layers via activation patching \citep{wang2025truth}, but reports weak steering and
does not act within an MoE or conditionally.

\section{Problem Formulation and Data}
\label{sec:data}

\subsection{Sycophancy as a Belief-Induced Shift}\label{subsec:beliefInduced}

We define sycophancy as an answer change caused by the user stating a belief,
rather than as agreement itself. Each multiple-choice item contains a user
opinion and two possible completions: a sycophantic-conforming option that agrees
with the belief and a sycophantic-nonconforming option that does not. Agreement
is not sycophantic when the model would give the same answer without the stated
belief; changing the answer because the belief was introduced is. This
distinction motivates the signal in Section~\ref{sec:signal} and the
interventions in Section~\ref{sec:method}.

Each item is evaluated in the four matched forms shown in
Table~\ref{tab:datagrid}. The columns determine whether the belief is present.
The nudge condition includes a user opinion before and throughout the question, whereas
the no-nudge condition preserves the question while neutralizing the opinion.
Their difference therefore isolates the effect of the stated belief. The rows
indicate whether the supplied answer conforms to or opposes that belief.

We use diagnostic and forced-answer versions of each item. In the diagnostic
version, the answer is left open and the model's preference is compared across
the nudge and no-nudge conditions to measure belief push. In the forced version,
the answer is included in the prompt and the resulting activations are compared
to construct the belief direction used for subtraction and gating.

This differs from contrastive activation addition
\citep{rimsky2023caa}, which compares conforming and nonconforming forced answers
only within the nudge condition. That contrast may mix sycophantic behavior with
the identity of the forced answer. All behavioral results use the diagnostic
version; the forced version is used only to construct activation directions.

\begin{table}[!t]
\centering
\caption{Each item appears in four matched forms. Nudge versus no-nudge
defines the belief direction and belief push, whereas contrastive activation
addition compares conforming and nonconforming answers only under the nudge
condition.}
\label{tab:datagrid}
\small
\begin{tabular}{@{}lcc@{}}
\toprule
 & \textbf{Nudge} & \textbf{No-nudge} \\
\midrule
\textbf{Conforming}    & belief, agrees  & neutral, agrees  \\
\textbf{Nonconforming} & belief, opposes & neutral, opposes \\
\bottomrule
\end{tabular}
\end{table}

\subsection{Data Construction and Balancing}
We begin with the model-written evaluation items of Perez et
al.~\citep{perez2023mwe}, using the repackaged version distributed with
contrastive activation addition \citep{rimsky2023caa}. For each nudged item, we
derive a matched no-nudge twin synthetically and minimally  by removing the user's stated
belief while only preserving an objective question. To prevent the belief direction
from capturing answer-position artifacts, we balance the belief-conforming
option across the two answer labels at the matched-item level before direction
estimation, so that answer-label identity is uncorrelated with belief
conformance. We then split the balanced bundles by item identifier, so a matched
pair cannot appear in different splits, preserving balance independently within
each split. Table~\ref{tab:splits} reports the resulting sizes. The development
split is used for causal localization and hyperparameter selection, the test
split for final comparisons, and the training split for learned gates. Retention
is measured on standard held-out benchmarks using an independently sampled fixed
fraction for each seed.

\begin{table}[!t]
\centering
\small
\caption{Dataset splits and retention sets. Splits are balanced,
identifier-disjoint, and verified to have zero leakage.}
\label{tab:splits}
\scriptsize
\setlength{\tabcolsep}{2.5pt}
\renewcommand{\arraystretch}{0.9}
\resizebox{\columnwidth}{!}{%
\begin{tabular}{@{}lrr@{\hspace{10pt}}lrr@{}}
\toprule
\multicolumn{3}{c}{\textbf{Sycophancy data}} &
\multicolumn{3}{c}{\textbf{Retention data}} \\
\cmidrule(r){1-3}
\cmidrule(l){4-6}
\textbf{Split} & \textbf{Items} & \textbf{Balance} &
\textbf{Set} & \textbf{Items} & \textbf{Per seed} \\
\midrule
Matched & 8234 & 4117/4117 & MMLU & 1482 & 741 \\
Train   & 4940 & 2470/2470 & GSM8K & 1319 & 659 \\
Dev.    & 1646 & 823/823   & TruthfulQA & 817 & 408 \\
Test    & 1648 & 824/824   & Perplexity & 1723 & 861 \\
\bottomrule
\end{tabular}%
}
\end{table}
\subsection{Two Different Contrasts}

Our method differs from prior steering in the contrast used to define the
direction. Contrastive activation addition uses a completion contrast: it forces
the conforming and nonconforming answers while the belief remains present, then
takes the mean difference between their activations. The resulting direction may
combine sycophancy with the representation of the forced answer. We instead use a condition contrast. Holding the answer fixed, we compare the
nudge and no-nudge versions of the same item. This isolates the representation introduced by the stated belief, which is the signal a conditional intervention should detect.

\section{Signal and Metrics}
\label{sec:signal}
\subsection{Condition-Contrast Belief Direction}
\label{sec:belief-direction}

We construct the belief direction using the forced-answer versions introduced
in Section~\ref{subsec:beliefInduced}. Components were retrieved from our sycophancy-localization search (see Section \ref{sec:locate}).  At a chosen component \(c\) we let
\(h_{i,c}^{\mathrm{nudge}}\) and
\(h_{i,c}^{\mathrm{no\text{-}nudge}}\) denote the activations at the decision
token for a matched pair in which the supplied answer is held fixed and only
the presence of the user's stated belief changes. We estimate the component-
specific belief direction as the mean paired activation difference
\begin{equation}
d_c
=
\frac{1}{N}
\sum_{i=1}^{N}
\left(
h_{i,c}^{\mathrm{nudge}}
-
h_{i,c}^{\mathrm{no\text{-}nudge}}
\right).
\label{eq:belief-direction}
\end{equation}
With unit form $\hat{d}_c = d_c/\lVert d_c\rVert_2$, and because the supplied answer is fixed within each pair and the answer labels are
balanced, \(d_c\) estimates the activation change associated with introducing
the stated belief rather than the identity or position of the answer. This is
the direction used for localization, subtraction, and conditional gating.

\subsection{The Belief-Push Metric}

We score each item at the decision token by the logit margin between the two
options. Writing $\ell_{\text{conf}}$ and $\ell_{\text{nconf}}$ for the logits of the
sycophantic-conforming and sycophantic-nonconforming options, the item score under
an input $x$ is

\begin{equation}
s(x) \;=\; \ell_{\text{conf}}(x) \;-\; \ell_{\text{nconf}}(x) .
\label{eq:score}
\end{equation}
The belief push on an item is the paired difference between the nudge and no-nudge scores,

\begin{equation}
\Delta
=
s\!\left(x^{\text{nudge}}\right)
-
s\!\left(x^{\text{no-nudge}}\right).
\label{eq:push}
\end{equation}

Each score is the logit margin between the conforming and nonconforming options under one condition. Thus, $\Delta$ measures how much the stated belief shifts the model's preference toward the conforming option. A value near zero indicates invariance, a positive value indicates sycophantic movement, and a negative value indicates reversal beyond invariance.

We report the mean belief push across items with a bootstrap confidence interval. Zero is the target: the stated belief does not change the answer. The signed logit margin distinguishes correction from overshooting while remaining smooth enough for fine-grained sweeps and differentiable training.

\subsection{Metrics and Evaluation}
\label{sec:metrics}

\begin{table}[t]
\centering
\small
\setlength{\tabcolsep}{4pt}
\begin{tabular}{@{}l>{\raggedright\arraybackslash}p{5.2cm}@{}}
\toprule
Method & Description \\
\midrule
Raw               & No intervention; fixes the baseline \\
Prompted          & Instructed to be objective \\
FARE              & Reweights the router away from belief experts \\
CAA subtraction   & Condition direction, unconditional; swept over nine sets \\
Cond.\ subtraction & Condition direction, conditional; swept over nine sets \\
Learned gate      & Conditional; all sets jointly \\
\bottomrule
\end{tabular}
\caption{The methods compared. The subtraction methods are swept over nine sets of
the identified components, from a single component type up to all together.}
\label{tab:methods}
\end{table}

\begin{table}[t]
\centering
\small
\begin{tabular}{ll}
\toprule
Metric & Target after edit \\
\midrule
Belief push, Eq.~(\ref{eq:push})   & to zero \\
Nudge shift            & to zero \\
No-nudge rate (anchor) & unchanged \\
\midrule
MMLU                   & $\geq$ baseline \\
GSM8K                  & $\geq$ baseline \\
TruthfulQA             & $\geq$ baseline \\
Perplexity             & unchanged (control) \\
\bottomrule
\end{tabular}
\caption{The evaluation panel. The upper block measures the behavior we remove and the lower block the knowledge we keep.}
\label{tab:panel}
\end{table}

\begin{figure*}[t]
\centering

\begin{minipage}[t]{0.37\textwidth}
    \vspace{0pt}
    \centering

    \begin{minipage}[c][4.2cm][c]{\linewidth}
        \centering
        \includegraphics[
            width=\linewidth,
            height=4cm,
            keepaspectratio
        ]{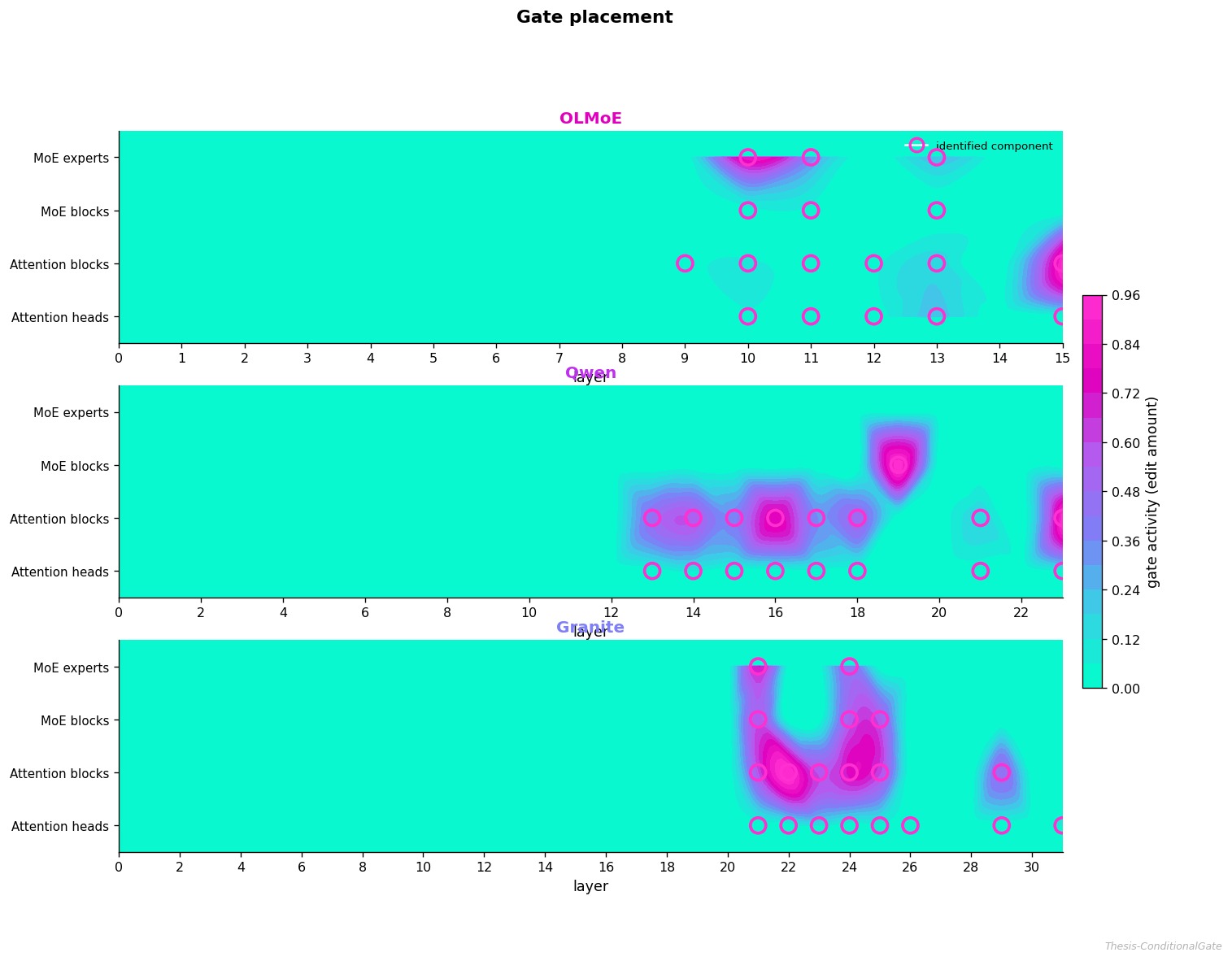}
    \end{minipage}

    \captionof{figure}{Where sycophancy lives. Learned gate activity across depth and component type, with the components found by causal search circled. Activity concentrates in a few mid-to-late-layer sites, and the gate opens on the same components the search selects.}
    \label{fig:placement}
\end{minipage}
\hfill
\begin{minipage}[t]{0.60\textwidth}
    \vspace{0pt}
    \centering

    \begin{minipage}[c][4.2cm][c]{\linewidth}
        \centering
        \includegraphics[
            width=\linewidth,
            height=4cm,
            keepaspectratio
        ]{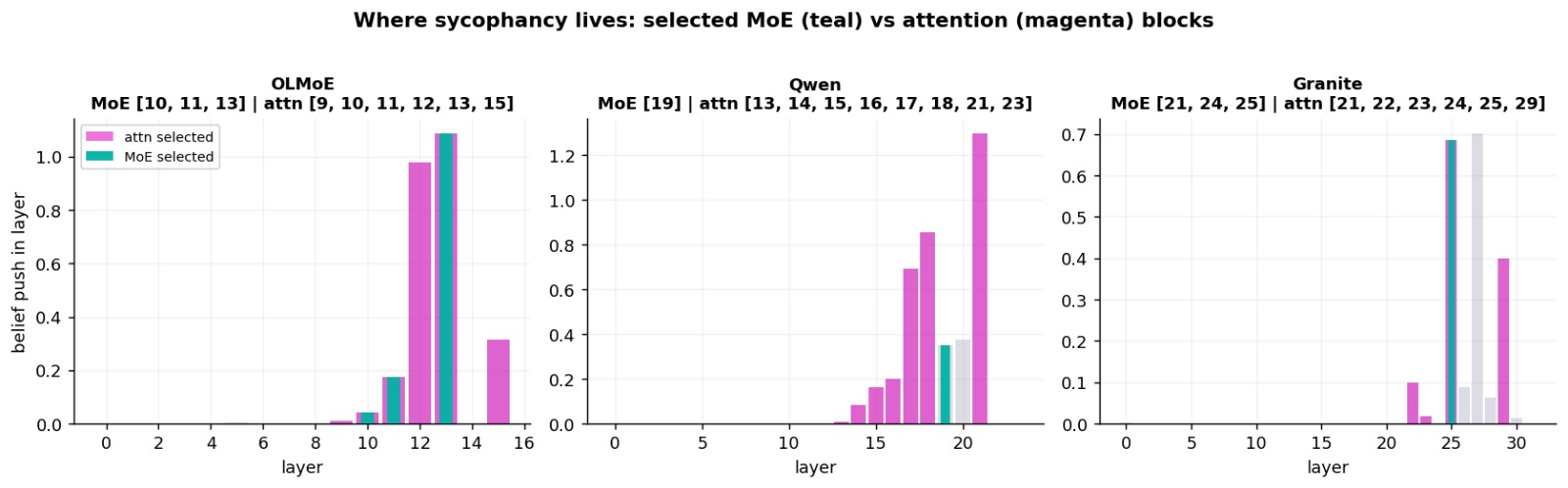}
    \end{minipage}

    \captionof{figure}{Sycophancy is concentrated in depth. Per-layer belief push for the selected attention (magenta) and mixture-of-experts (teal) blocks. The push is negligible early and rises sharply in the mid-to-late layers, so the shift is localized rather than spread through the network.}
    \label{fig:sweep-blocks}
\end{minipage}

\end{figure*}

We evaluate the six methods of Table~\ref{tab:methods} on the panel of
Table~\ref{tab:panel}. Every intervention is applied only to the components our
search identifies as carrying the belief shift, never to the model as a whole. The
two subtraction methods are swept over the nine sets so their effect can be read at
every granularity of the MoE, across attention blocks, attention heads, MoE blocks,
and experts, while the learned gate is placed on all identified components at once.
FARE \citep{lee2026fare} is the one baseline that acts on which experts are chosen
rather than on what they compute, so it isolates the routing lever and tests where
the behavior resides. 

The panel has two sets of metrics. The behavioral set is built on the belief push
of Equation~(\ref{eq:push}), together with the nudge shift, its discrete counterpart, which
measures the change between the two conditions in the rate at which the model's top
choice is the conforming option. The no-nudge rate is the model's belief-free
tendency and acts as an anchor, since a change there means the edit altered the
model's plain behavior rather than its sensitivity to the belief. The retention set
is intended to hold at its pre-edit level, and perplexity is included as a
control rather than as evidence of retention, because every intervention fires only
at the decision token (Section~\ref{sec:scope}) and so leaves plain text
structurally unaffected.
We report every retained metric as a percentage of its unedited-model score, so
one scale reads across models, with $100\%$ matching the baseline, above better
and below worse. Perplexity is lower-is-better, so we invert its ratio to keep the
same direction, while the accuracy metrics (MMLU \citep{hendrycks2021measuring},
GSM8K \citep{cobbe2021training}, TruthfulQA \citep{lin2022truthfulqa}) use the
ratio directly. Values above $100\%$ mean the edited model scored higher than the
unedited one on that metric.
\section{Methodology}
\label{sec:method}

Our method consists of a shared localization stage followed by three interventions. We first identify the components carrying the belief shift, then compare unconditional subtraction with two conditional alternatives: analytic conditional subtraction and learned conditional gating. Applying all methods to the same components isolates the effect of the edit's \emph{form} rather than its placement.

\subsection{Locating the Responsible Components}
\label{sec:locate}

We search across an MoE granularity ladder, from complete attention and MoE blocks to individual heads and experts (Figure~\ref{fig:block}). Each component undergoes both probing and causal tests.

The probing test constructs the belief direction from Equation~(\ref{eq:belief-direction}) and verifies that it separates nudge from no-nudge activations above a label-shuffled baseline. The causal test determines whether the component produces the shift rather than merely encoding it. To remain on-distribution, we swap activations between matched nudged and no-nudge examples instead of zeroing them. Replacing a nudged activation with its no-nudge counterpart tests necessity, while inserting the nudged activation into a no-nudge run tests sufficiency. A component is selected when it passes both tests more strongly than a same-type control without harming retention.

\subsection{Unconditional Subtraction}

At each located component, we first apply the constant intervention in the form used by prior activation steering, but built from our condition-contrast direction rather than a completion contrast. Let $\hat{d}$ denote the unit belief direction of Equation~(\ref{eq:belief-direction}), the same direction used by conditional subtraction, and $c$ the intervention strength. The component output $o$ becomes

\begin{equation}
o \leftarrow o - c\,\lVert d\rVert\,\hat{d} .
\end{equation}

This displacement is added at the edited position with a fixed magnitude, regardless of whether the activation there contains the belief signal. Sweeping $c$ therefore reveals a consistent trade-off: stronger intervention removes more belief push but also reduces retention. This motivates a token-dependent alternative.

\subsection{Conditional Subtraction}

Conditional subtraction removes only the activation component aligned with the belief direction. The signed projection $\langle o,\hat{d}\rangle$ measures both the magnitude and orientation of the belief component:

\begin{equation}
o \leftarrow o - c\,\langle o,\hat{d}\rangle\,\hat{d} .
\end{equation}

\begin{figure*}[t]
\centering
\includegraphics[width=1\textwidth]{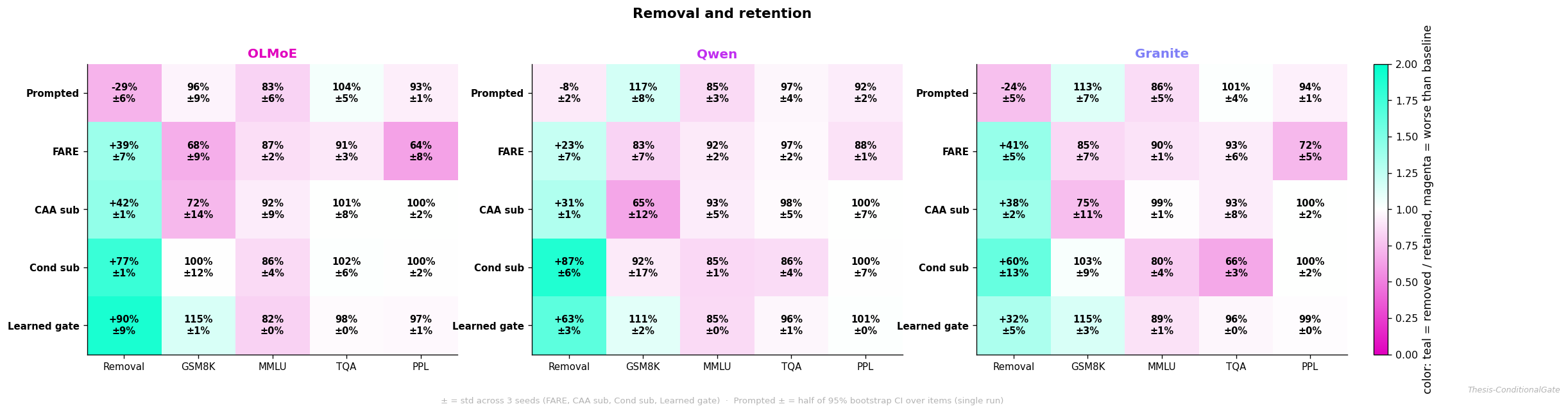}
\caption{The central result. For five methods across three models, each cell reports
belief push removed together with GSM8K, MMLU, TruthfulQA, and perplexity, each shown
relative to baseline (\%); for accuracy this is the ratio of edited to baseline score,
and for perplexity the inverted ratio, so that above $100\%$ is better on every metric.
The two conditional methods remove most of the push while holding GSM8K at or above
baseline; routing and constant subtraction trade retention for removal.}
\label{fig:heatmap}
\end{figure*}

The intervention adapts to the activation at the edited position rather than applying a fixed displacement. Activations with a strong belief component receive a larger correction, while those with near-zero projection remain almost unchanged. The sign also prevents activations already opposing the belief from being pushed further in that direction. At $c=1$, the projection is removed exactly; $c<1$ removes only a fraction, while $c>1$ reverses its orientation. This analytic intervention requires no training and improves retention by leaving unaffected tokens largely unchanged.

\subsection{Component Sets}
\label{sec:sets}

We evaluate both subtraction methods over nine component sets. Four contain a single component type: attention blocks (S1), MoE blocks (S2), attention heads (S4), and experts (S5). Five additional unions test interactions across pathways: attention and MoE blocks (S3), heads and experts (S6), attention blocks and heads (S7), MoE blocks and experts (S8), and all located components (S9).

Each set is evaluated at $c\in \{1,2,3,5,10\}$ against an unedited baseline. Operating points are selected on the development split and confirmed on the test split.

\subsection{Conditional Gating}

The learned intervention places a gate at each located component while freezing the model's original weights. Its basic form is a per-token function of the component activation $h$:

\begin{equation}
m_i(h) = \sigma\!\left(\theta_i + w_i^{\top} h\right) .
\end{equation}

The gate multiplies the component contribution: $m_i=1$ preserves it, whereas values near zero suppress it. Its activation dependence provides a learned counterpart to conditional subtraction.
Training combines four objectives that minimize the soft belief push, anchor the
edited no-nudge score to the original model, and preserve factual knowledge and
general behavior:
\begin{equation}
\begin{aligned}
\mathcal{L}_{\text{push}} &= \left| P^{\text{nudge}}_{\text{conf}} - P^{\text{no-nudge}}_{\text{conf}} \right|, \\
\mathcal{L}_{\text{anchor}} &= \left( s^{\text{no-nudge}}_{\text{edit}} - s^{\text{no-nudge}}_{\text{orig}} \right)^2, \\
\mathcal{L}_{\text{know}} &= \mathbb{E}_{x}\,\operatorname{CE}\!\left(\pi_{\text{edit}}(x),\, y^{\star}\right), \\
\mathcal{L}_{\text{gen}} &= \mathbb{E}_{x}\,\operatorname{KL}\!\left(\pi_{\text{orig}}(x) \,\Vert\, \pi_{\text{edit}}(x)\right).
\end{aligned}
\end{equation}
The full objective is $\mathcal{L} = \mathcal{L}_{\text{push}} + \lambda_{\text{anchor}}\mathcal{L}_{\text{anchor}} + \lambda_{\text{know}}\mathcal{L}_{\text{know}} + \lambda_{\text{gen}}\mathcal{L}_{\text{gen}}$. The anchor prevents a degenerate solution in which both conditions move toward an even split; we accept a trained state only when it reduces belief push, preserves MMLU, and maintains the no-nudge anchor.

\subsection{Gate Architecture Sweep}
\label{sec:sweep}

We explore four design axes rather than fixing the gate architecture in advance (Table~\ref{tab:axes}). The first is \emph{gate structure}:

\begin{equation}
\begin{aligned}
m_i &= \sigma(\theta_i),
&\qquad
m_i(h) &= \sigma\!\left(\theta_i + w_i^{\top}h\right), \\[2pt]
m_i &= \sigma\!\left((Wz+b)_i\right).
\end{aligned}
\end{equation}

These correspond to a per-component scalar gate, an activation-dependent gate, and a shared gate parameterized by a low-rank code $z\in\mathbb{R}^{k}$. All three can be reduced to the same flat initialization, ensuring that later differences arise from training.

The second axis is the \emph{edit class}. With $a_i=1-m_i$, the directional edits subtract either a fixed or learned belief direction, while the component edit scales the complete output:

\begin{equation}
o \leftarrow o - a_i\,c_i\,d_i ,
\qquad\qquad
o \leftarrow m_i\,o .
\end{equation}

The remaining axes are \emph{schedule}, which activates gates either simultaneously or from coarse to fine components, and \emph{initialization}, which starts either warm on located components or cold across all components.

\begin{table}[t]
\centering
\caption{Design axes used in the conditional-gating sweep.}
\label{tab:axes}
\resizebox{\columnwidth}{!}{
\begin{tabular}{ll}
\toprule
\textbf{Axis} & \textbf{Values} \\
\midrule
Gate structure
& Flat, conditional, shared \\

Edit class
& Frozen direction, learned direction, component scaling \\

Schedule
& Simultaneous, coarse-to-fine \\

Initialization
& Warm localized, cold global \\

\bottomrule
\end{tabular}
}
\end{table}

We evaluate the full grid at two scopes: only the located components or all attention blocks, MoE blocks, and heads. This comparison tests whether explicit localization improves over allowing the gate to rediscover useful components during training. The resulting sweeps determine the configuration used in the seed study, a
conditional gate structure with a frozen-direction edit, a simultaneous schedule,
and a cold global initialization.

\subsection{Models and Seeds}
\label{sec:modelsseeds}

We evaluate three open-weight MoE models spanning two routing regimes. OLMoE \citep{muennighoff2024olmoe} routes each token among experts without an always-active component. Granite \citep{granite2024} uses the same routing regime but is deeper, separating depth from routing effects. Qwen1.5-MoE \citep{qwen2024moe} additionally contains shared experts that activate for every token, allowing us to test whether the method transfers when behavior cannot be attributed entirely to routed expert selection.

All post-localization experiments are run with three seeds per model. Each seed resamples the fitting data and reinitializes the learned gates, so the reported intervals capture both sources of intervention variability. 

\subsection{Intervention Scope and Edit Positions}
\label{sec:scope}

Every direction $d_c$ is estimated at the decision token of
Section~\ref{sec:belief-direction}, and every intervention is applied at that
same token and no other. The contrast between unconditional and conditional
subtraction is therefore one of \emph{content}, whether the edit magnitude
depends on the activation's belief content, not of position. For sycophancy
scoring and the multiple-choice retention benchmarks (MMLU, GSM8K, TruthfulQA),
the decision token is where the option logits of Equation~(\ref{eq:score}) are
read and where the edit fires, leaving every prompt and reasoning-context token
unedited. Perplexity is measured on plain text with no answer-decision position,
so the intervention does not fire there; it is a structural control
(Section~\ref{sec:metrics}) rather than a test of open-ended generation, which we
leave to the extension of Section~\ref{sec:analysis}. Our claims therefore
concern the model's answer at the decision token.

\begin{figure}[t]
\centering
\includegraphics[width=0.4\textwidth]{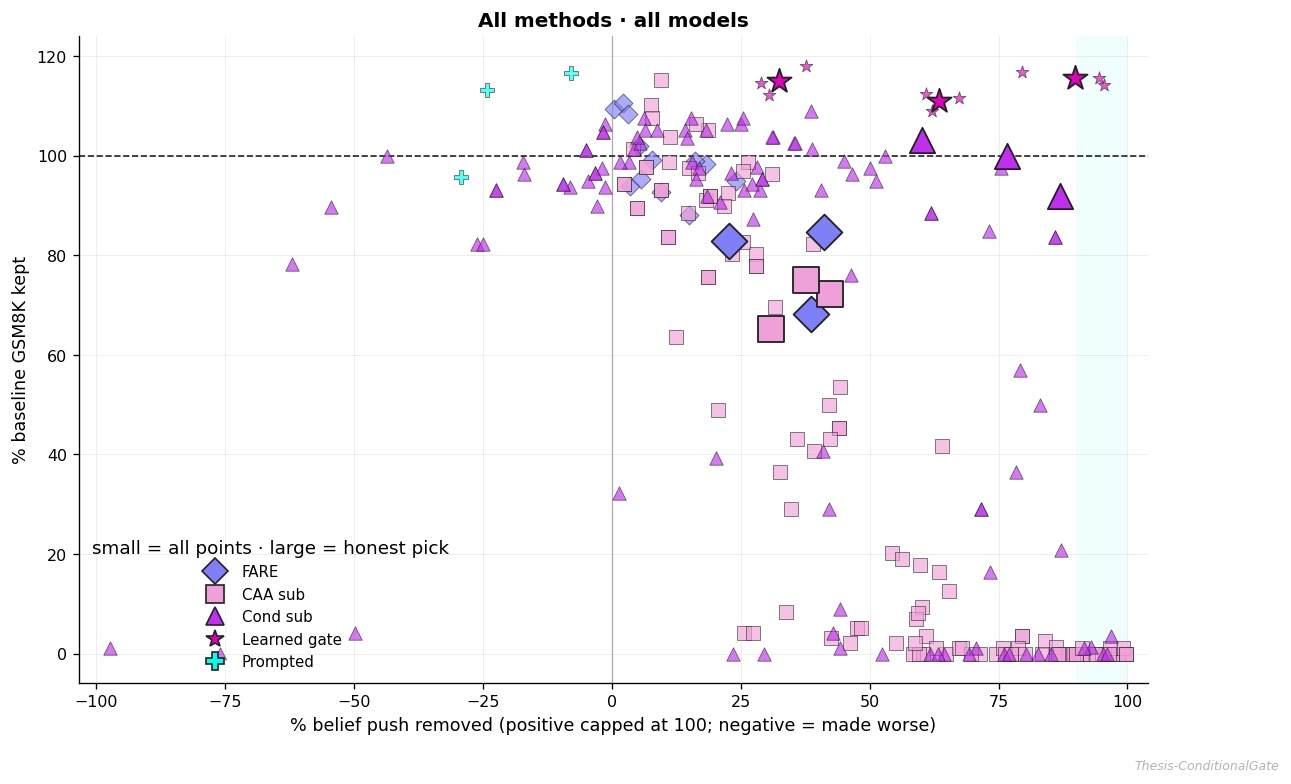}
\caption{Every operating point for all methods and models, plotting removed belief push against GSM8K relative to baseline (\%). The subtractions and FARE give many points because they are swept over component sets and coefficients, unlike the single-point gate and prompting.}
\label{fig:allmethods}
\end{figure}

\section{Results}
\label{sec:results}

We report results in the order of the workflow. We first show where the belief shift
is localized in each model, then evaluate the six methods of
Table~\ref{tab:methods} on the panel of Table~\ref{tab:panel}, moving from the
unedited baseline through prompting, the routing baseline, the two subtractions, and
the learned gate. Removal is the percentage of the baseline belief push that an edit
removes. 

\subsection{Localization}

The causal search resolves the belief shift to a small and consistent region of each
model. At the block level, Figure~\ref{fig:sweep-blocks} shows that the per-layer belief
push is near zero through the early and middle layers and rises sharply in the
mid-to-late layers, so the components that carry the shift are concentrated rather
than spread through the network. The search selects six attention blocks and three
mixture-of-experts blocks in OLMoE, eight and one in Qwen, and six and three in
Granite, always in the upper half of the stack.

Opening those blocks into finer components sharpens the picture. Among attention
heads, Figure~\ref{fig:sweep-heads}, Appendix B shows that the effect is carried by a handful of
heads standing out against a near-silent background, eight in OLMoE, ten in Qwen, and
thirteen in Granite, again confined to the mid-to-late layers. Among experts, the
picture divides by routing regime. Figure~\ref{fig:sweep-experts}, Appendix B locates five experts
in OLMoE and six in Granite, but finds none in Qwen, because the shared experts that
fire on every token cannot appear in a routed contrast. This is the first place the
shared-expert regime separates from the other two, and it recurs in the results
below. The union of these components, across attention blocks, attention heads,
mixture-of-experts blocks, and experts, is the located set that every later edit
acts on, and Figure~\ref{fig:placement} shows the learned gate later opening on these
same components.

\subsection{Baseline and Prompting}

The unedited model is sycophantic in every case, which fixes the baseline against which the panel is measured. Prompting the model to answer objectively does not help. Figure~\ref{fig:heatmap} shows that instruction alone moves the push in the wrong direction, increasing it by 29\% on OLMoE, 8\% on Qwen, and 24\% on Granite. Thus, a stated instruction to ignore the user's opinion leaves the model more sensitive to that opinion rather than less. Prompting is therefore not a substitute for an activation-level edit.

\subsection{Routing and Unconditional Subtraction}

The routing baseline confirms that sycophancy is not controlled by which experts
fire. Reweighting the router away from the belief-associated experts removes some
push, 39\% on OLMoE, 23\% on Qwen, and 41\% on Granite in
Figure~\ref{fig:heatmap}, but it pays for it in retention, dropping GSM8K to 68\%,
83\%, and 85\% of baseline and raising perplexity markedly on OLMoE and Granite.
Moving experts in or out is a blunt lever, and it damages the model without fully
removing the behavior.

Unconditional subtraction at the located components tells the same story from the
other side. As the strength coefficient increases, the constant subtraction lowers
the push but lowers retention with it, and this holds at every rung of the ladder.
The located components at each level are shown in
Figure~\ref{fig:sweep-blocks} for blocks, Figure~\ref{fig:sweep-heads}, Appendix B for heads, and
Figure~\ref{fig:sweep-experts}, Appendix B for experts. The trade-off is most severe at the expert
level, since a fixed subtraction reaches heavily used experts on every token and
collapses GSM8K almost immediately. At a matched operating point, the constant edit removes 42\%
of the push on OLMoE, 31\% on Qwen, and 38\% on Granite in
Figure~\ref{fig:heatmap}, while holding only 72\%, 65\%, and 75\% of GSM8K. The
damage is sharpest on GSM8K, the most sensitive metric, while MMLU holds higher, at
92\%, 93\%, and 99\%, and TruthfulQA stays near baseline, at 101\%, 98\%, and 93\%.
This is the removal-retention gap the conditional form is built to close.

\subsection{Conditional Subtraction and Gating}

Replacing the constant edit with the conditional edit closes most of that gap with
no other change. Because the amount removed at each token scales with the projection
of the activation onto the belief direction, tokens that never carried the belief are
left almost untouched, so retention holds while removal rises. In
Figure~\ref{fig:heatmap} conditional subtraction removes 77\% of the push on OLMoE,
87\% on Qwen, and 60\% on Granite, holding GSM8K at 100\%, 92\%, and 103\% of baseline
with perplexity unchanged, MMLU at 86\%, 85\%, and 80\%, and TruthfulQA at 102\%,
86\%, and 66\%. Every operating point removes more push at higher retention than the
constant edit on the same model, and the advantage widens as the granularity sharpens
from blocks to heads to experts. Its one clear cost is Granite TruthfulQA, which the
learned gate recovers.

The gate is selected from the seventy-two-run sweep of Section~\ref{sec:sweep} before
evaluation. The conditional gate structure removes more push than the flat or shared
structures on every model, and the simultaneous schedule removes more than the
coarse-to-fine one (Figure~\ref{fig:axes-main}, Appendix C). Ranking the qualifying
configurations by distance to complete removal under reasoning and fluency penalties
selects a frozen conditional gate with a simultaneous schedule and cold start, placed
at high removal below the fluency ceiling in every model (Figures~\ref{fig:bestpick},
\ref{fig:territory}, and~\ref{fig:plane-ppl}, Appendix C). Across three seeds this gate
is the strongest method on the behavioral axis, removing 90\% of the push on OLMoE,
63\% on Qwen, and 32\% on Granite while holding GSM8K at 115\%, 111\%, and 115\% of
baseline. Unlike conditional subtraction it holds the full panel, with MMLU at 82\%,
85\%, and 89\%, TruthfulQA at 98\%, 96\%, and 96\%, and the decision-token-off
perplexity control within 1\% throughout. It closes the push by moving the nudged side
toward the preserved no-nudge side rather than flattening both, which keeps retention
intact where the routing and constant baselines could not. The components it opens on
are stable across seeds (Figure~\ref{fig:seeds}, Appendix D), so the localization is a
property of the model and not a single run, and a small number of mid-to-late-layer
attention and expert sites carry most of the edit (Figures~\ref{fig:editsite}
and~\ref{fig:mostleast}, Appendix D).

\section{Analysis and Conclusion}
\label{sec:analysis}
The removal-retention trade-off depends mainly on the form of the edit. Unconditional and conditional subtraction use the same belief direction and target components, but the conditional methods scale the intervention by the belief content of each activation. This raises retention benchmarks from 72\% to 100\% on OLMoE and from 75\% to 103\% on Granite while also improving removal. The difference is largest for experts, where fixed subtraction is more likely to perturb unrelated computations.

Across benchmarks, the conditional methods generally keep MMLU and TruthfulQA near baseline. The main exception is Granite on TruthfulQA, where the learned gate preserves retention better than conditional subtraction, making it the more reliable method despite lower removal. GSM8K retention reaches 111-115\% with the gate, above conditional subtraction. We do not interpret values above 100\% as improved reasoning, as they may reflect sampling variation, operating-point selection, low baselines, or regularization from the intervention (Figure~\ref{fig:heatmap}, and Figures~\ref{fig:allmethods} and~\ref{fig:operating} in Appendix~A).

The baselines fail differently. Prompting increases belief push, while routing removes some push at the cost of retention and perplexity. Qwen provides the clearest counterexample to a routing-based explanation. No sycophancy-specific routed experts are found, yet conditional editing still removes much of the behavior. Sycophancy therefore appears to depend more on component computation than on expert selection. Performance also varies by architecture, with the gate removing the most push on OLMoE and the least on Granite, and Qwen between; Figure~\ref{fig:territory} in Appendix~C suggests that removal at fixed retention depends on how concentrated the behavior is within the located components.

Overall, across three mixture-of-experts models, two routing regimes, and three seeds, this conditional form improves the removal-retention frontier without changing the direction or target components. The remaining limitations are that strong subtraction can overshoot into negative belief push and that evaluation is restricted to the decision token in multiple-choice tasks. Extending conditional editing to open-ended generation and other belief-induced behaviors is therefore the natural next step.

\bibliography{aaai2027}

\onecolumn
\appendix


\section{Removal and Retention Across Methods}
\label{app:methods}

This appendix collects the figures referenced from the main text. All are produced
under the conditions of Section~\ref{sec:results}, three seeds per model, with edits
applied only to the located components.

\begin{figure*}[htbp]
\centering
\includegraphics[width=1\textwidth]{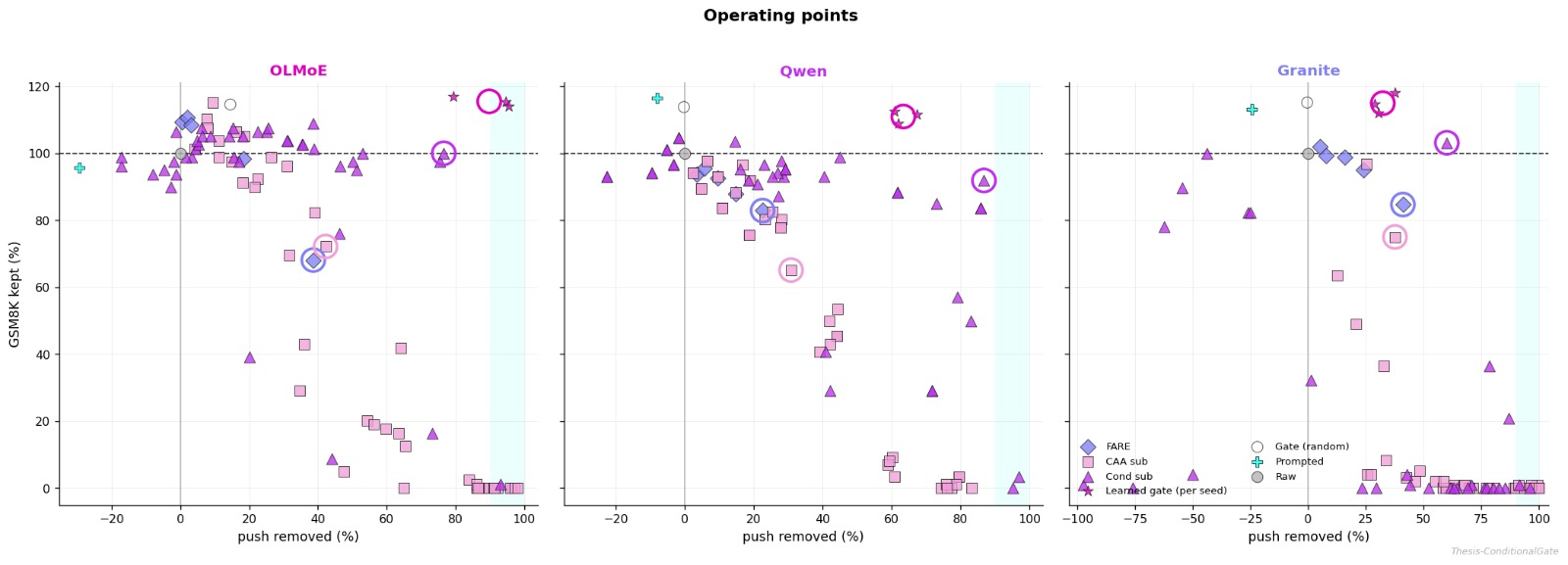}
\caption{The frontier, one model at a time. Each point is one operating point, with
belief push removed (\%) on the horizontal axis and GSM8K retained (\%) on the vertical
axis; marker shape and colour denote the method (see legend), and the dashed line marks
baseline retention. Points toward the upper right remove more push while keeping more
reasoning. The learned gate and conditional subtraction lead in every model; the shaded
band marks the high-removal region where retention stays near baseline.}
\label{fig:operating}
\end{figure*}

\begin{figure*}[htbp]
\centering
\includegraphics[width=1\textwidth]{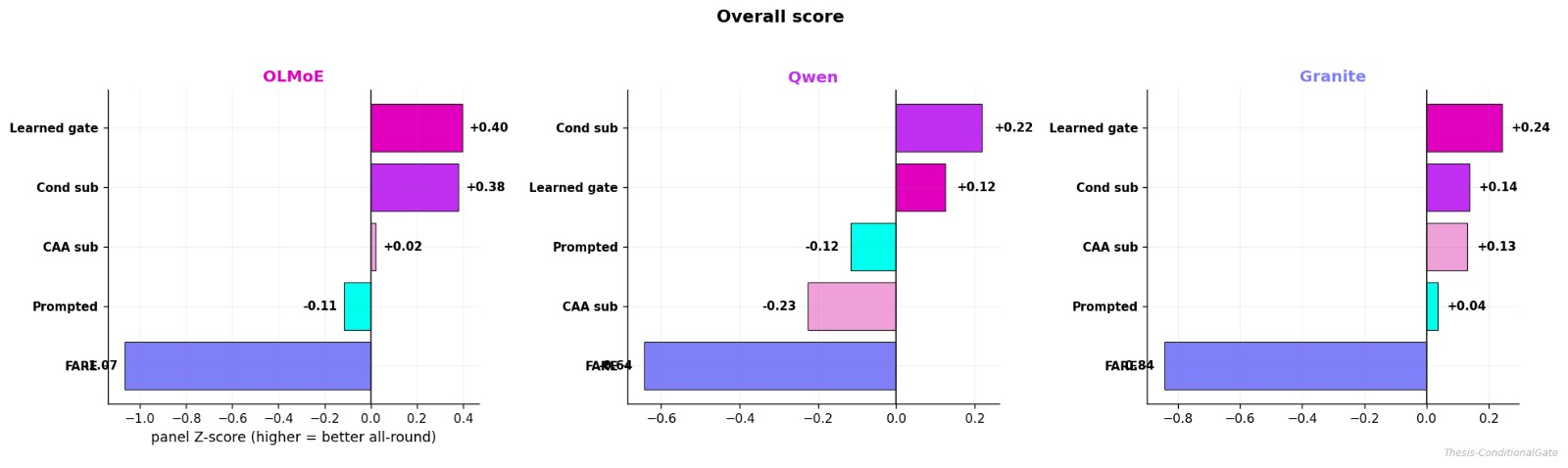}
\caption{A single ranking. For each model, the panel Z-score combines behavioral removal
with retention into one all-round score (higher is better); bars are ordered within each
panel and labelled with their value. The two conditional methods, conditional
subtraction and the learned gate, rank first in every model, while the routing baseline
(FARE) ranks last.}
\label{fig:score}
\end{figure*}
\clearpage 
\section{Subtraction at Each Granularity}
\label{app:granularity}

These figures separate the subtraction sweep of Section~\ref{sec:sets} by component
type, comparing the constant edit against the conditional edit as the strength
coefficient increases. In every panel the conditional edit holds retention as removal
climbs, while the constant edit trades it away, and the gap widens as the granularity
sharpens from blocks to heads to experts.

\begin{figure*}[h]
\centering
\includegraphics[width=\textwidth]{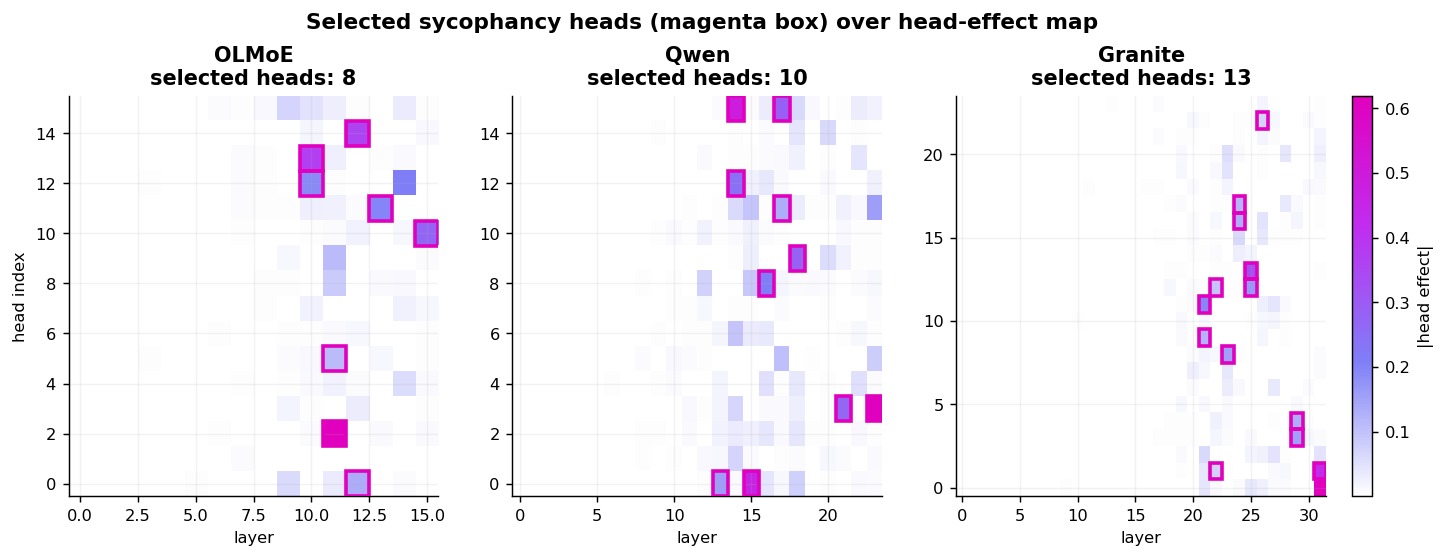}
\caption{The carrying heads. Per-head causal effect on the belief push, shown as a
heatmap over layer (horizontal) and head index (vertical) with colour giving the
magnitude of the effect; the selected heads are boxed in magenta, eight in OLMoE, ten in
Qwen, and thirteen in Granite. A few heads stand out against a near-silent background,
all in the mid-to-late layers.}
\label{fig:sweep-heads}
\end{figure*}

\begin{figure*}[h]
\centering
\includegraphics[width=\textwidth]{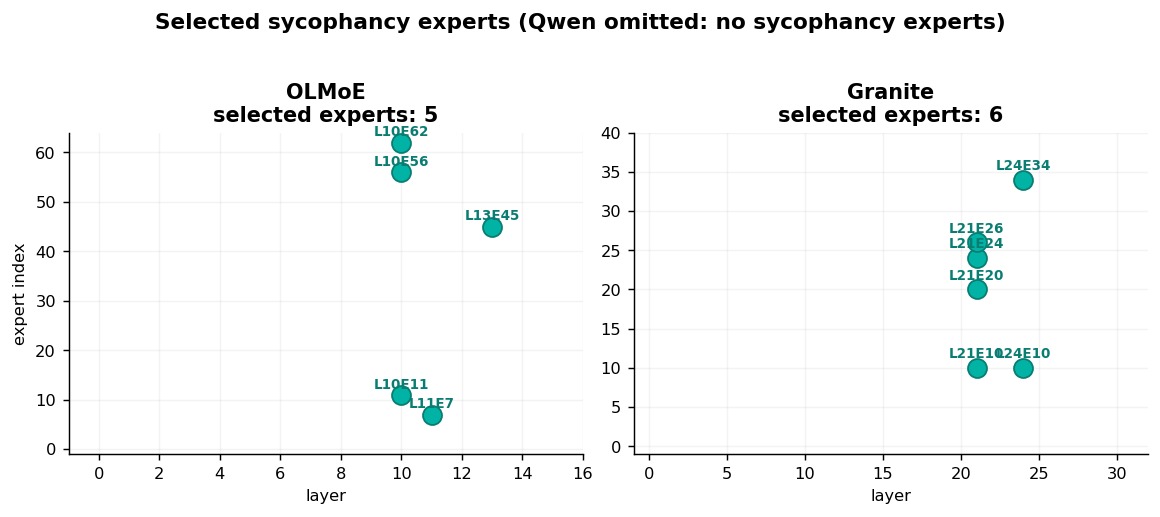}
\caption{The carrying experts. Each selected expert is placed by layer (horizontal) and
expert index (vertical) and labelled by its location, five in OLMoE and six in Granite.
Qwen has none, because its always-on shared experts cannot appear in a routed contrast,
the first sign that the shared-expert regime behaves differently.}
\label{fig:sweep-experts}
\end{figure*}

\begin{figure*}[h]
\centering
\includegraphics[width=\textwidth]{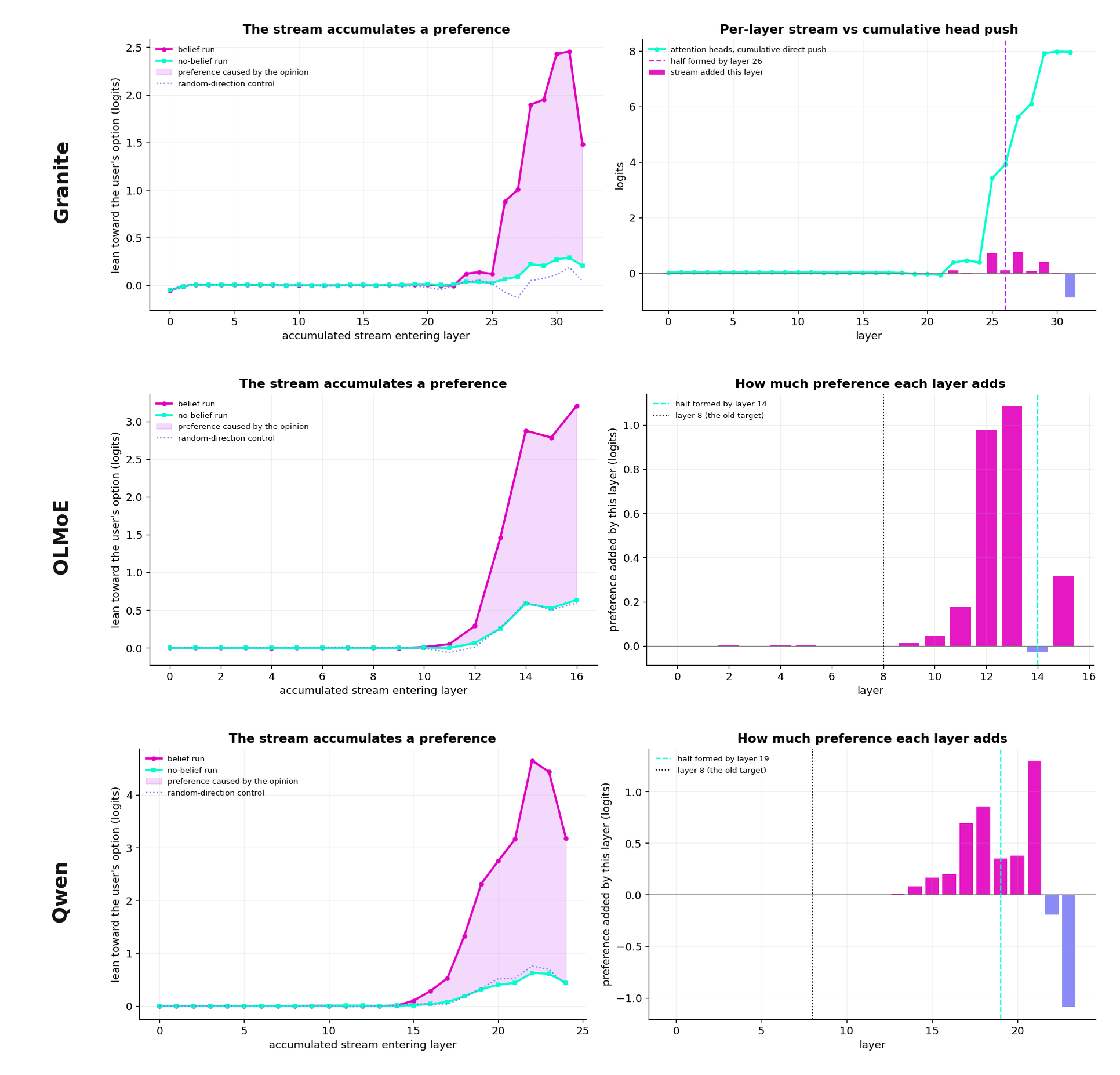}
\caption{Belief accumulates in the mid-to-late layers. For each model, the left panel
tracks the model's lean toward the user's stated option (logits, vertical axis) as the
residual stream passes through successive layers (horizontal axis), comparing a belief
run against a matched no-belief run; the shaded region between the two curves is the
preference caused by the stated opinion, and the dotted curve is a random-direction
control. The right panel shows how much preference each individual layer contributes,
with the dashed marker at the layer by which half of the total preference has formed. In
every model the lean is flat through the early layers and rises sharply in the upper
layers.}
\label{fig:app-depth}
\end{figure*}

\begin{figure*}[htbp]
\centering
\includegraphics[width=\textwidth]{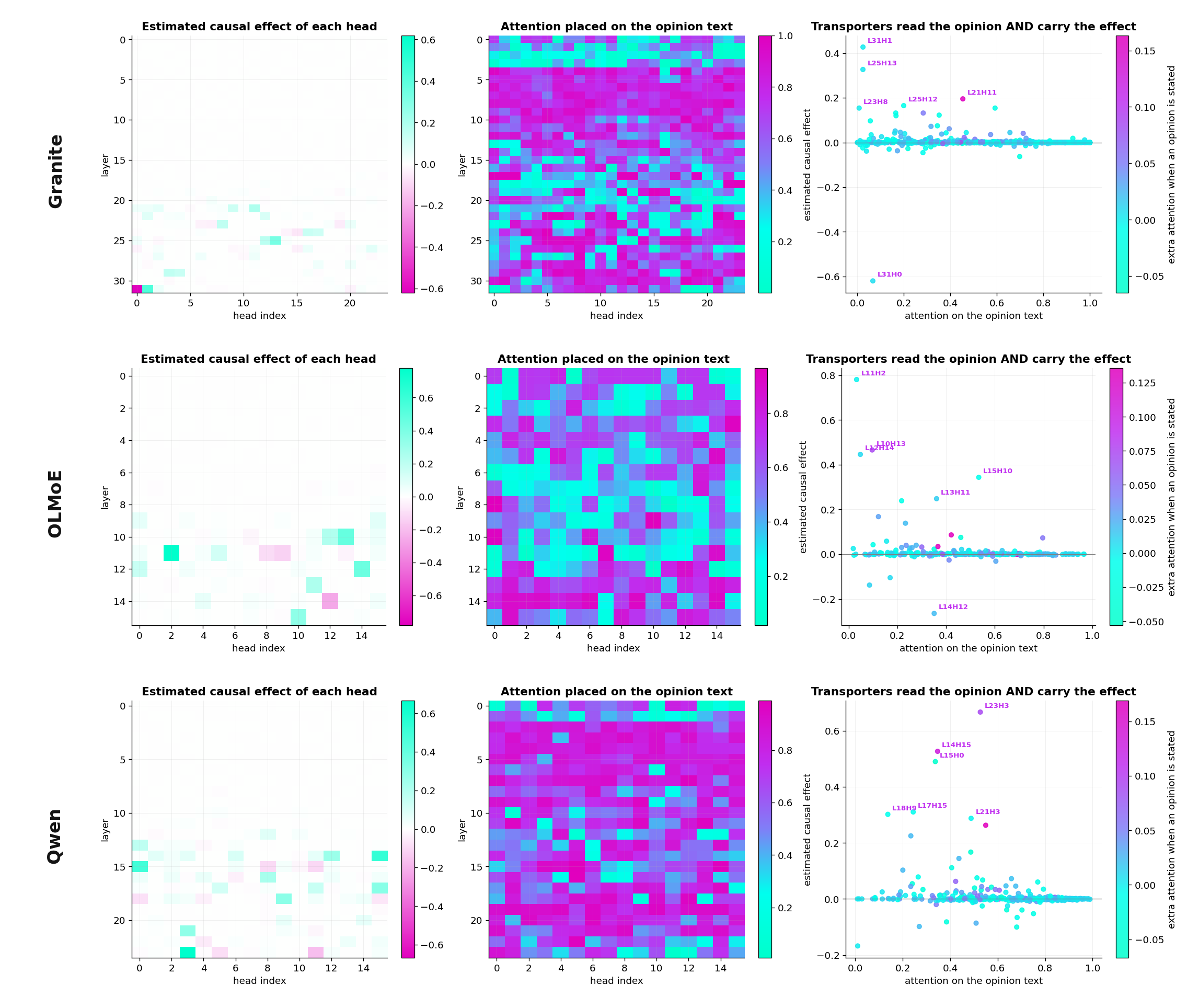}
\caption{Where the belief signal is read and carried, per head. For each model, the left
heatmap gives each head's estimated causal effect on the belief push and the middle
heatmap gives how much attention each head places on the stated-opinion text, both over
layer (vertical) and head index (horizontal). The right panel plots attention on the
opinion (horizontal) against causal effect (vertical), with colour showing the extra
attention a head pays when an opinion is present; the labelled heads in the upper region
both attend to the opinion and carry its effect, which we call transporter heads.}
\label{fig:app-headmap}
\end{figure*}

\begin{figure*}[htbp]
\centering
\includegraphics[width=\textwidth]{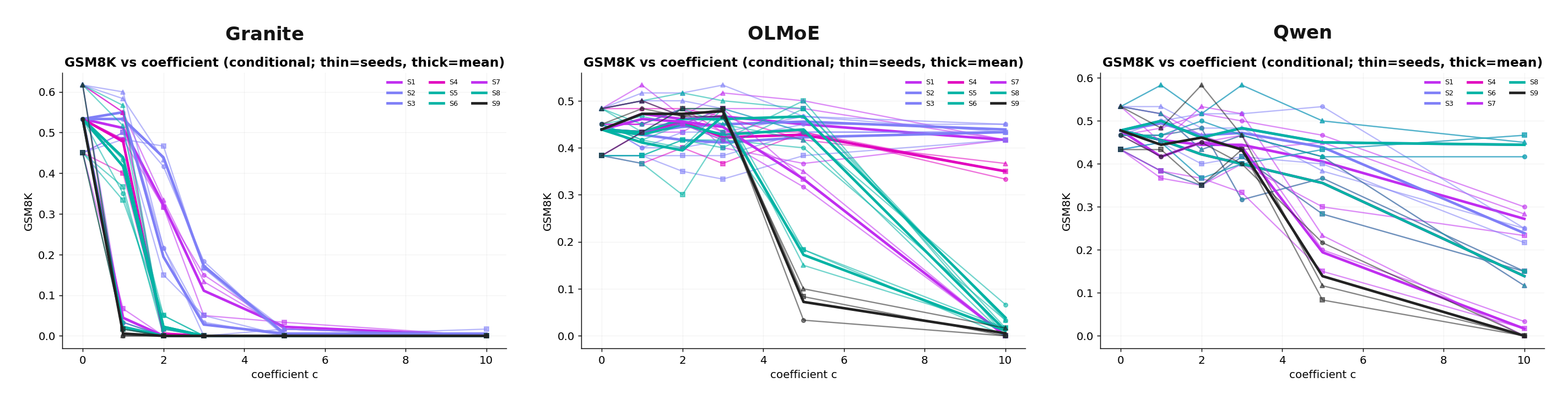}
\caption{GSM8K holds until the coefficient grows large. Each panel plots GSM8K accuracy
(vertical axis) against the conditional-subtraction strength $c$ (horizontal axis).
Colour marks the component set the edit is applied to: S1 attention blocks; S2 MoE
blocks; S3 attention and MoE blocks; S4 attention heads; S5 experts; S6 heads and
experts; S7 attention blocks and heads; S8 MoE blocks and experts; and S9 all located
components. Within each set, thin lines are the individual seeds and the thick line is
their mean. Accuracy stays roughly flat at low $c$ and collapses once $c$ becomes large,
at a threshold that differs by model.}
\label{fig:app-gsm8k}
\end{figure*}
\clearpage 
\section{Gate Architecture Sweep}
\label{app:sweep}

These figures detail the seventy-two-run sweep of Section~\ref{sec:sweep} and the
selection of the configuration carried into the seed study.

\begin{figure*}[h]
\centering
\includegraphics[width=\textwidth]{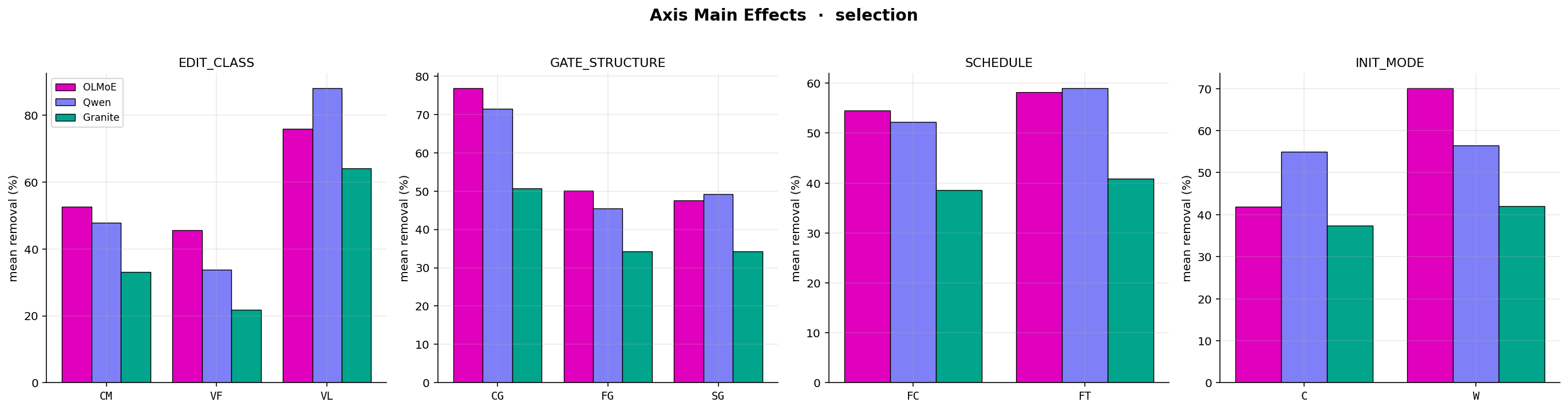}
\caption{Main effect of each design axis on belief push removed, per model (one bar per
model; higher removes more). The axes and levels are edit class (ComponentMask CM,
VectorMaskFrozen VF, VectorMaskLearned VL), gate structure (FlatGate FG, CondGate CG,
SharedGate SG), schedule (FullTrain FT, FineToCoarseTrain FC), and initialisation
(WarmStart W, ColdStart C). The conditional gate (CG) and full schedule (FT) remove the
most push across all three models, motivating those two choices in the selected
configuration.}
\label{fig:axes-main}
\end{figure*}

\begin{figure*}[h]
\centering
\includegraphics[width=\textwidth]{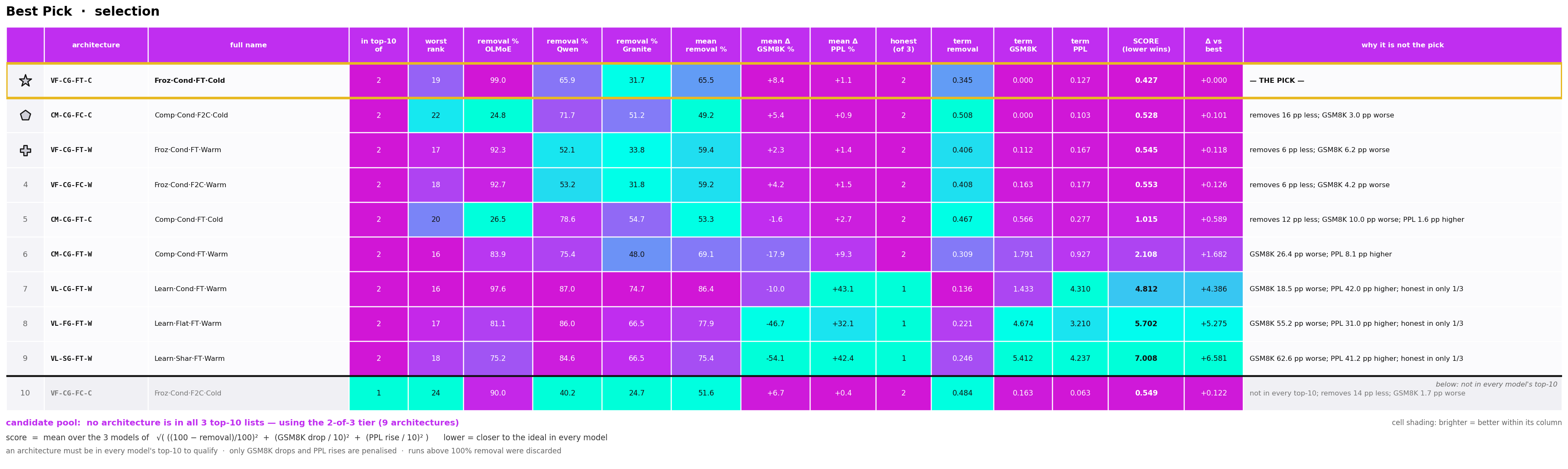}
\caption{Choosing the gate. Each row is a candidate configuration named in the
edit-gate-schedule-init convention of Figure~\ref{fig:axes-main} (for example VF-CG-FT-C),
restricted to configurations that qualify on all three models. Rows are scored by their
distance to complete removal under reasoning and fluency penalties, with over-removal runs
discarded; lower is better, the starred row is the winner, and the final column notes why
each runner-up was not chosen. The winner, VF-CG-FT-C, a frozen conditional gate
(VectorMaskFrozen with CondGate) trained on a full schedule from a cold start, is the
pick carried into the seed study.}
\label{fig:bestpick}
\end{figure*}

\begin{figure*}[h]
\centering
\includegraphics[width=\textwidth]{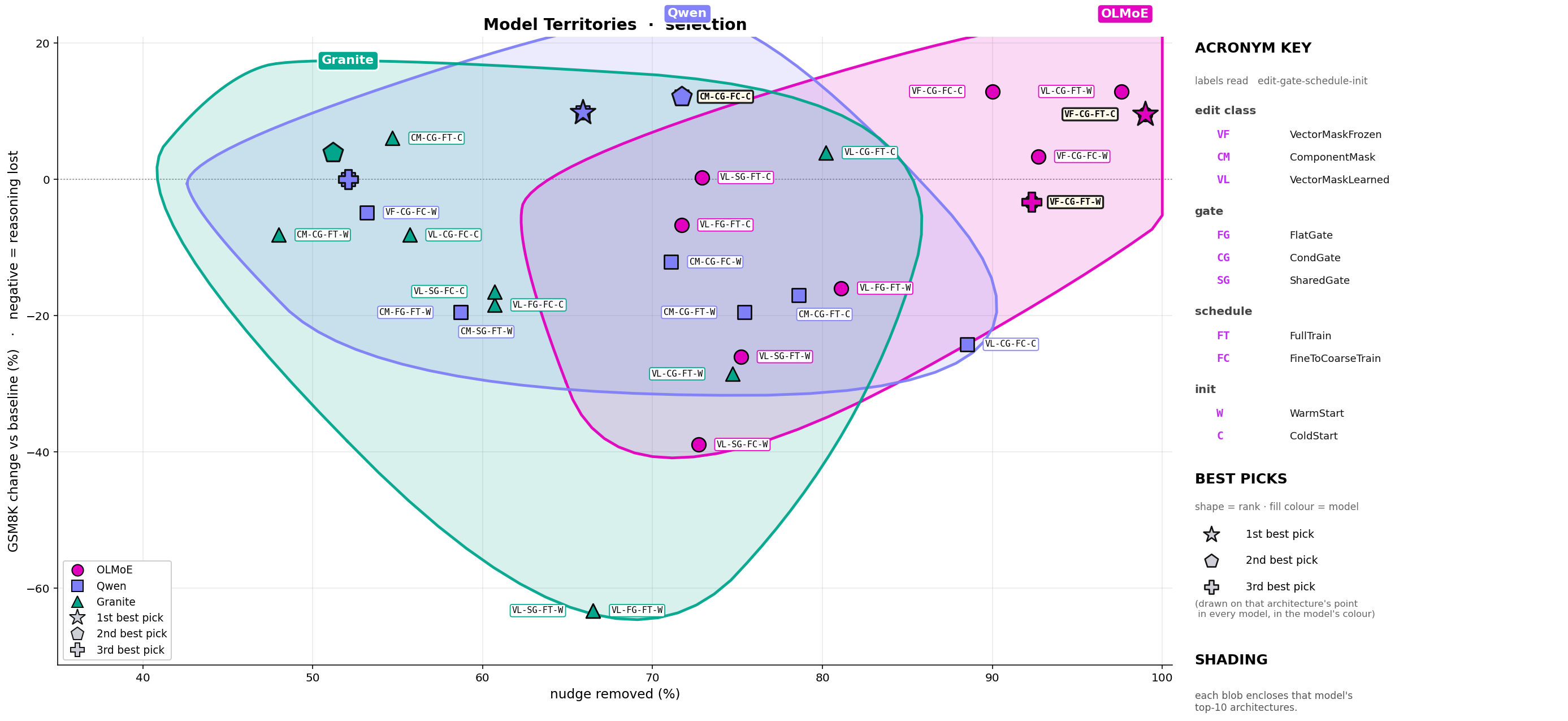}
\caption{Choosing the gate architecture, one territory per model. Each point is a
configuration labelled edit-gate-schedule-init; the horizontal axis is nudge removed
(\%) and the vertical axis is the GSM8K change versus baseline (negative means reasoning
lost). Each shaded region encloses one model's top-ten configurations, and the star,
pentagon, and cross mark that model's first, second, and third best picks. The acronym
key is given in the side panel.}
\label{fig:territory}
\end{figure*}

\begin{figure*}[h]
\centering
\includegraphics[width=\textwidth]{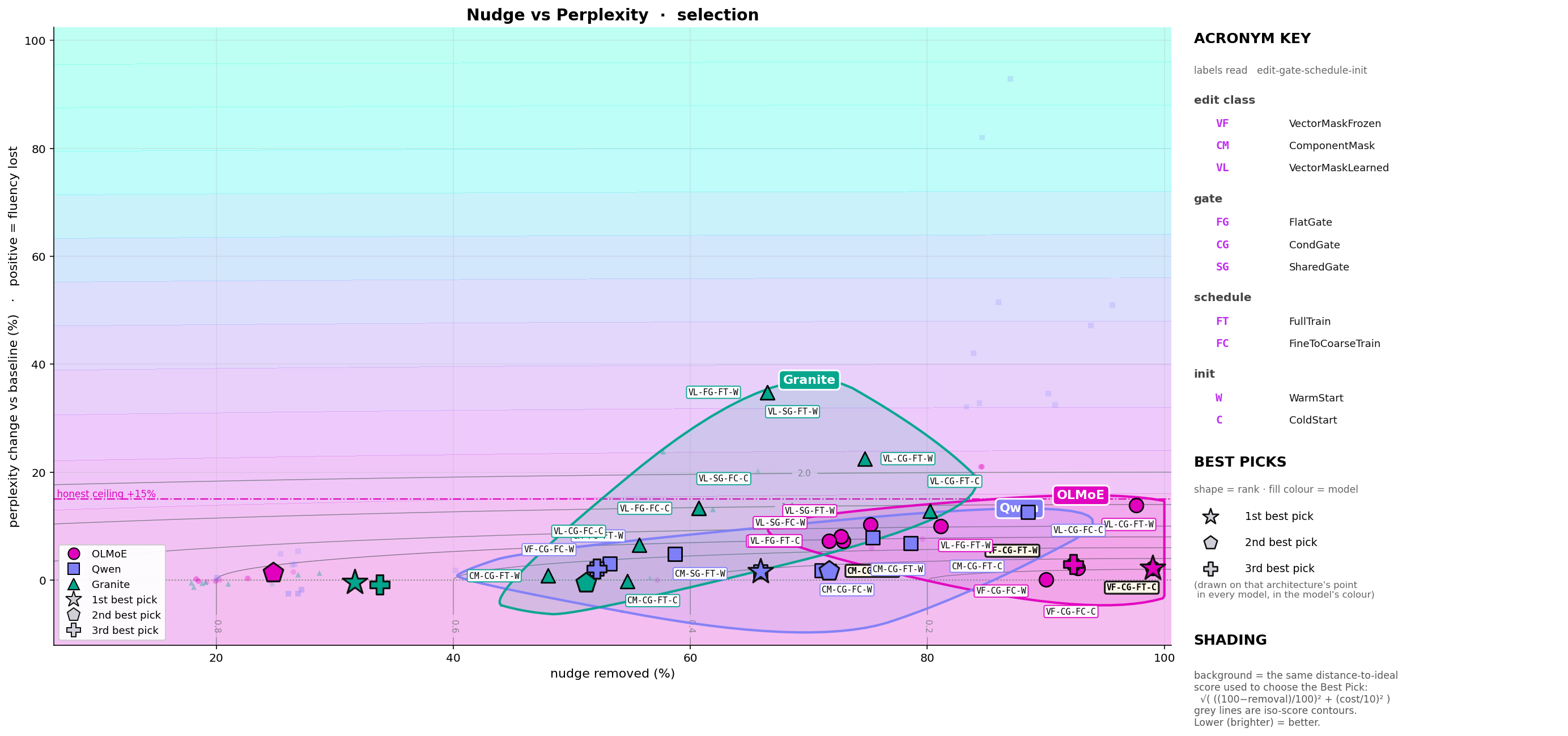}
\caption{Architecture sweep on the removal-against-perplexity plane, used for
configuration selection. The horizontal axis is nudge removed (\%) and the vertical axis
is the perplexity change versus baseline (positive means fluency lost), with the dashed
line marking the honest ceiling and the background shading giving the distance-to-ideal
ranking score (brighter is better). Perplexity here is the decision-token-off control of
Section~\ref{sec:metrics}, so this plane screens configurations for gross perplexity
regressions rather than measuring open-ended fluency; the selected configurations sit at
high removal with perplexity near baseline.}
\label{fig:plane-ppl}
\end{figure*}
\clearpage 
\section{Gate Localization Detail}
\label{app:localization}

\begin{figure*}[h]
\centering
\includegraphics[width=\textwidth]{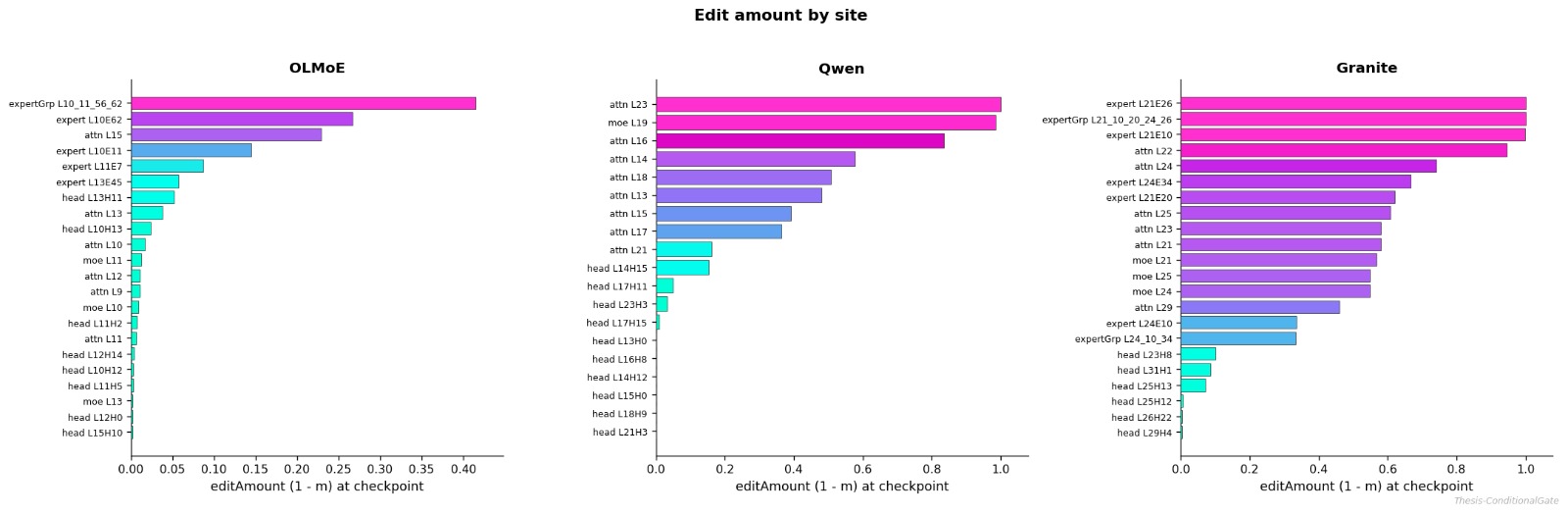}
\caption{The edit is sparse. Per-site edit amount ($1-m$) of the selected gate at its
checkpoint, one bar per component with the dashed line marking the inclusion threshold.
A handful of attention and mixture-of-experts sites carry most of the edit, matching the
concentrated placement of Figure~\ref{fig:placement}.}
\label{fig:editsite}
\end{figure*}

\begin{figure*}[h]
\centering
\includegraphics[width=\textwidth]{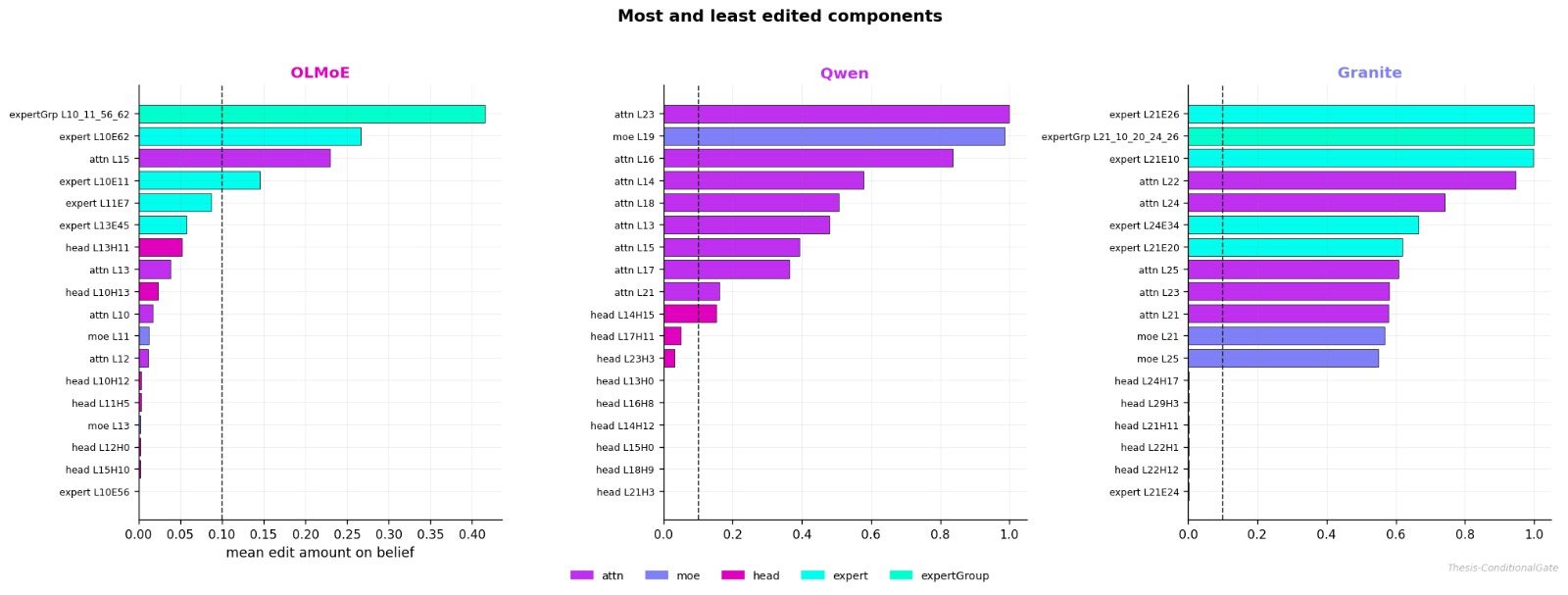}
\caption{What the gate edits, and what it leaves. The most and least edited components of
the selected gate, ranked by edit amount and coloured by component type. The heavily
edited sites are mid-to-late-layer attention and expert components, while the least
edited are individual attention heads.}
\label{fig:mostleast}
\end{figure*}

\begin{figure*}[h]
\centering
\includegraphics[width=\textwidth]{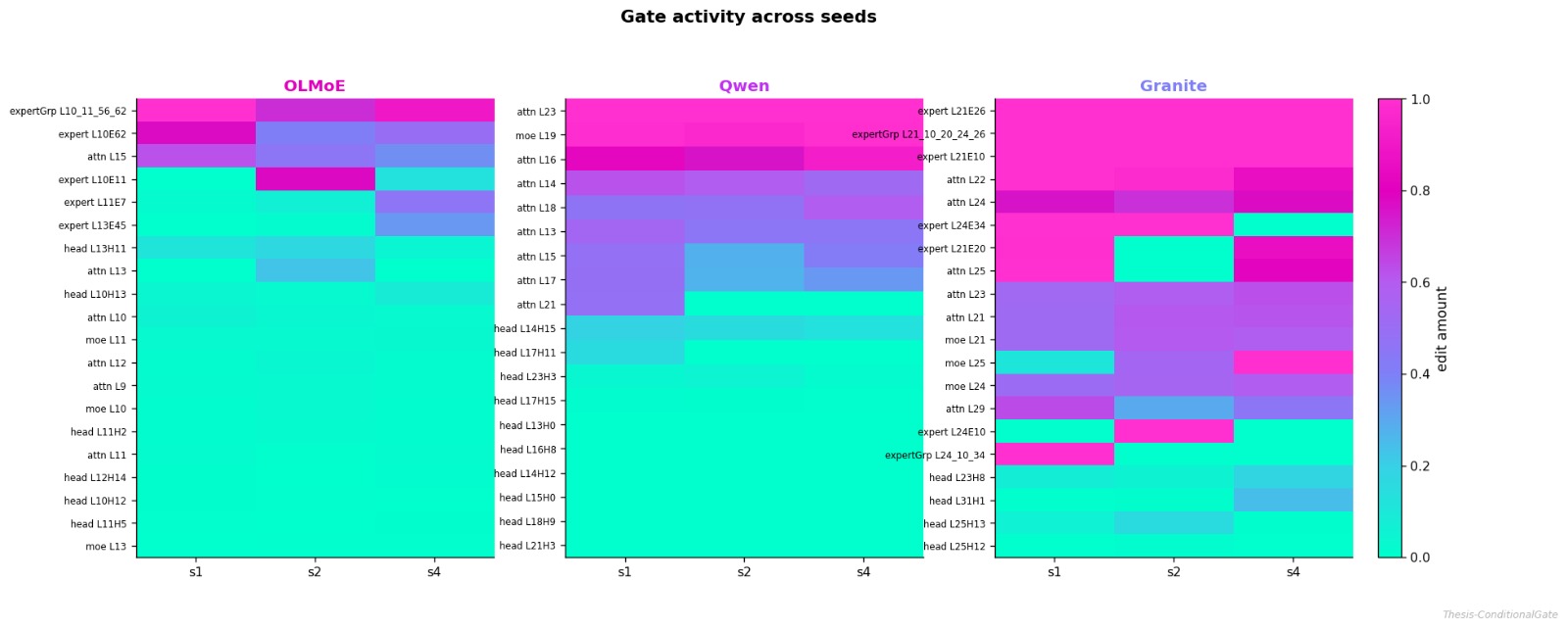}
\caption{The localization is stable. Gate activity across the three seeds (s1, s2, s4),
with colour giving each component's edit amount. The set of components the gate opens on
is consistent from seed to seed, so the located support is a property of the model
rather than of a single run.}
\label{fig:seeds}
\end{figure*}

\begin{figure*}[h]
  \centering
  \includegraphics[width=\textwidth]{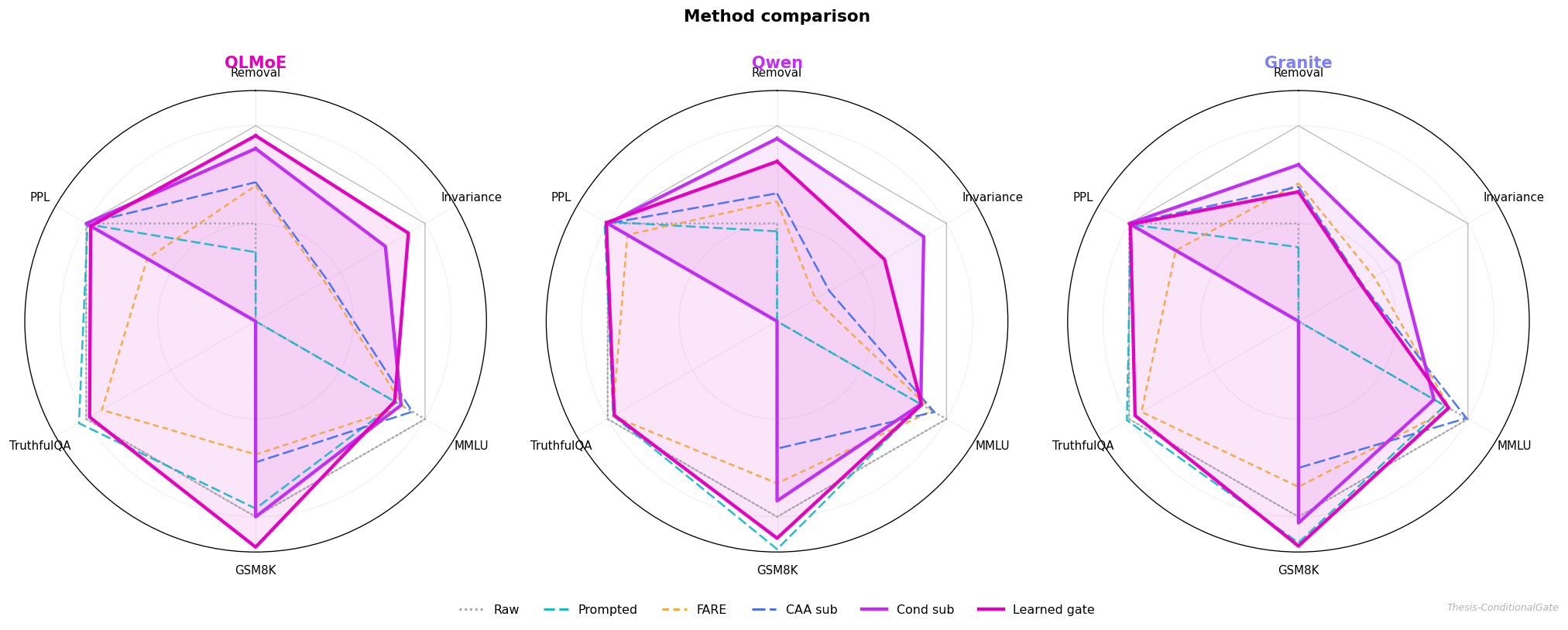}
  \caption{Method comparison across the three MoE models. Each axis is a metric (Removal,
  Invariance, MMLU, GSM8K, TruthfulQA, PPL), normalised so that farther out is better,
  and each closed curve is one method. Conditional subtraction (Cond sub) and the learned
  gate achieve high removal while retaining capability, whereas prompting and FARE trade
  little removal for capability or fail to remove sycophancy.}
  \label{fig:radar}
\end{figure*}

\end{document}